\documentclass{article}
\usepackage{arxiv}
\usepackage{multirow} 
\usepackage{algorithm}
\usepackage{algorithmic}
\usepackage{amsmath} 
\usepackage{tcolorbox}
\usepackage{booktabs}
\usepackage{rusty}
\usepackage{enumitem}
\setlist[itemize]
{align=parleft,left=0pt..1em}
\usepackage{hyperref}

\newcommand{\sln}{\texttt{UniTranslator}}

\begin{document}

\title{Collaboration is all you need: LLM Assisted Safe Code Translation}

\author{ \href{https://orcid.org/0000-0000-0000-0000}{\includegraphics[scale=0.06]{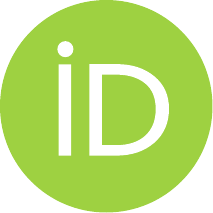}\hspace{1mm}Rabimba Karanjai} \\
	University Of Houston\\
	\texttt{rkaranjai@uh.edu} \\
	\And
	{Sam Blackshear} \\
	Mysten Labs\\
	\texttt{sam@mystenlabs.com} \\
    	\And
	{Lei Xu} \\
	Kent State University\\
	\texttt{xuleimath@gmail.com} \\
    	\And
	{Weidong Shi} \\
	University Of Houston\\
	\texttt{wshi3@Central.UH.EDU} \\
}

\maketitle
\begin{abstract}
 This paper introduces \sln{}, a visionary framework that reimagines code translation as a collaborative endeavor among multiple, compact LLMs.  By orchestrating the interaction of specialized agents, each focused on different aspects of the translation process and grounded in a deep understanding of programming concepts, \sln{} achieves a level of accuracy and efficiency that rivals larger, monolithic models.  Our preliminary evaluation demonstrates the potential of \sln{} to overcome the limitations of existing approaches and unlock the power of smaller LLMs for complex code translation tasks. We explore the effectiveness of this dynamic multi-agent paradigm in handling diverse language pairs, including low-resource languages, and in mitigating common issues such as code artifacts and hallucinations through the use of Natural Language Inference (NLI) grounding and iterative feedback mechanisms. 
\end{abstract}


\section{Introduction}
\label{sec:intro}

Code translation, the process of converting source code from one programming language to another while preserving its functionality, is becoming increasingly critical in our rapidly evolving software landscape. From modernizing legacy systems \cite{krishna2021transforming,perez2021software,kaleem2021event} to enabling seamless cloud migration \cite{nitin2022cargo}, the ability to automatically and accurately translate code is paramount. While large language models (LLMs) have shown promise in this domain \cite{jana2023attention,roziere2020unsupervised,karanjai2024teaching}, their effectiveness is often hampered by model size, computational cost, and a narrow focus on specific language pairs \cite{xia2024understanding,pan2024lost}.

We envision a future where code translation is universally accessible, efficient, and accurate, regardless of the source and target languages or the available computational resources. To realize this vision, we introduce \sln{}, a groundbreaking framework that leverages the power of collaboration among multiple, compact LLMs.  \sln{} reimagines code translation as a multi-agent endeavor, where specialized LLMs, each with unique expertise, work together to achieve high-fidelity translations and proactively address potential bugs. This approach not only transcends the limitations of monolithic models but also democratizes access to advanced code translation capabilities by enabling deployment on commonly available hardware.


This research introduces a dynamic, multi-agent framework for code translation, leveraging a heterogeneous ensemble of Large Language Models (LLMs) orchestrated by a Director LLM. The system features a Code Translation Pipeline tailored to specific language pairs and tasks.  Key components include: (1) an NLI-based Claim Checker for semantic fidelity; (2) a repository of benchmarked LLMs with language-specific performance profiles; (3) a dynamic prompt generation system; and (4) a compiler suite for real-time code validation. The Director LLM adaptively selects and coordinates LLMs and specialized agents (e.g., Code Explainer, Concept Verifier) based on continuous feedback and task requirements. This research investigates the feasibility of this adaptive, multi-agent architecture for achieving high-fidelity and efficient code translation. Which takes us to the following research questions:

\begin{itemize}
    \item \textbf{Harnessing Collective Intelligence:} Can we unlock the synergistic potential of smaller LLMs by orchestrating their collaboration in a multi-agent framework, effectively creating a "virtual expert" that surpasses the capabilities of individual models?
    \item \textbf{Concept-Driven Translation:} Can we guide the translation process by explicitly incorporating conceptual knowledge of programming languages, leading to more accurate and semantically sound code transformations, even for low-resource languages?
    \item \textbf{Mitigating Errors and Hallucinations through Grounded Reasoning:}  Can we leverage carefully created Natural Language Inference (NLI) to ground the translation process in established coding practices, thereby minimizing the risk of introducing language-specific artifacts or generating nonsensical code?
    \item \textbf{Proactive Bug Remediation via Feedback Loops:} Can we create a self-improving system where compiler feedback and guided hints are used to refine translations bugs iteratively?
\end{itemize}

This paper presents our initial steps towards realizing this vision. We detail the architecture of \sln{}, highlighting its innovative multi-agent quorum, concept-guided translation mechanism, and feedback-driven bug reduction strategy. Preliminary evaluations demonstrate the potential of \sln{} to achieve accurate translations even with compact LLMs, paving the way for a future where code flows seamlessly between languages, fostering greater interoperability and accelerating software innovation.

\section{The \sln{} Architecture: A Symphony of Collaborating LLMs}
\label{sec:arch}

The core of our vision lies in a novel architecture that orchestrates a collaborative ensemble of LLMs, each contributing unique strengths to the code translation process.  \sln{} is not merely a collection of models; it is a carefully designed ecosystem where compact LLMs, acting as specialized agents, interact and learn from each other, guided by a shared understanding of programming concepts.

\subsection{Grounded in Knowledge, Driven by Concepts}

At the foundation of \sln{} lies a robust knowledge grounding mechanism. We leverage a specialized Natural Language Inference (NLI) model, enhanced for code-specific tasks, to filter out potential "code hallucinations" – instances where the translation process might introduce code snippets that are syntactically incorrect or inconsistent with the target language's paradigms. This NLI model acts as a guardian of code fidelity, ensuring that the generated code adheres to established programming practices.

We implemented fact checking on programming and coding languages using NLI by leveraging the capabilities of NLI models to validate the accuracy of statements or claims related to programming languages and their features. This approach determines whether a given statement about a coding language is true, false, or uncertain based on established knowledge about that language. We train our NLI following 
Honovich, Or, et al. ~\cite{honovich2022true}, but on a dataset, we created for each programming language based on its documentation from the official documentation.

Furthermore, we introduce a novel "concept agent" – an LLM fine-tuned on a corpus of programming language textbooks and documentation. This agent embodies a deep understanding of fundamental programming concepts, transcending the limitations of surface-level syntax. By incorporating this conceptual knowledge, \sln{} can generate translations that are not only syntactically correct but also semantically sound, capturing the true essence of the original code.

\subsection{A Quorum of Experts}

\sln{}'s multi-agent framework is built upon a "quorum" of LLMs, each playing a distinct role in the translation process. A "Director" LLM intelligently assesses the complexity of the input code, determining the appropriate level of agent involvement. For simpler translations between common languages, a single, powerful code LLM might suffice. However, for more challenging scenarios, such as translating to or from low-resource languages or across vastly different programming paradigms, the Director activates a collaborative team of agents.

This team includes specialized code agents, each proficient in a particular programming language, and the aforementioned concept agent. These agents engage in a dynamic exchange, guided by the Director, to produce the most accurate and idiomatic translation.  The selection of agents is not static; it is a fluid process, informed by few-shot prompting and a continuous assessment of each agent's contribution to the task. For instance, the Director might initially send the code to the concept agent to create a conceptual blueprint. This blueprint is then evaluated by the code agents, leading to revisions and refinements until a consensus is reached.

\subsubsection{Agent Selection: A Dynamic and Intelligent Process}

Our proposed framework assumes a discrete set of choices regarding programming paradigms, libraries, and feedback signals. Within this architecture, two pivotal entities are introduced: Small Expert LLMs, which function as autonomous agents responsible for selecting code implementations, and DirectorLLM, an entity that provides guidance based on observational data.

The DirectorLLM operates asynchronously, offering recommendations that inform the decision-making process of the Small Expert LLMs. This framework delineates a formalized interaction between automated decision-making mechanisms and intelligent advisory inputs. The primary objective is to enhance the coding process by effectively integrating individual expertise with collaborative insights, thereby optimizing both performance and accuracy in software development tasks.

If we formalize at time $t$ over state $s \in S$ with code selection observation $o \in O$ and selection suggestions $o^s \in A$ is 
$p(s_t \mid a_{0:t-1}, o_{0:t}, o^s_{0:t})$. Using Bayes’ theorem, we can rewrite our expression as:

\begin{align*}
& p(s_t | a_{0:t-1}, o_{0:t}, o^s_{0:t})  \\ 
& \propto p(o^s_t | s_t, a_{0:t-1}, o_{0:t-1}, o^s_{0:t-1}) \\
& \quad \cdot p(o_t | s_t, a_{0:t-1}, o_{0:t-1}, o^s_{0:t-1}) \\
& \quad \cdot p(s_t | a_{0:t-1}, o_{0:t-1}, o^s_{0:t-1}) \qquad (1) 
\end{align*}

where the subscript $0:t$ refers to all instances of that variable from $0$ to $t$, and $t^- = t - 1$. This expression can be simplified using the independence assumption, the law of total probability, and the Markov property to

\begin{align*}
& p(s_t | a_{0:t-1}, o_{0:t}, o^s_{0:t})  \\
& \propto p(o^s_t | s_t) p(o_t | s_t, a_{t-1}) \\
& \quad \cdot \sum_{s_{t-} \in S} p(s_t | s_{t-}, a_{t-1}) p(s_{t-} | a_{t-1}, o_{t-1}, o^s_{t-1}). \qquad (2)
\end{align*} 

For challenging translations, the Director initiates a collaborative process. The source code is first analyzed by the concept agent, which extracts the underlying programming concepts.  These concepts are then used to guide the selection and interaction of specialized code agents, ensuring that the translation process is informed by a deep understanding of the code's semantic meaning.  This dynamic selection and collaboration process allows \sln{} to adapt to a wide range of translation scenarios, effectively leveraging the collective intelligence of the agent team. The agents then engage in a multi-turn conversation, refining the translation through proposals and critiques until a consensus is reached, mimicking the collaborative efforts of human expert programmers.

\subsection{Iterative Refinement through Feedback}

\sln{} incorporates a powerful feedback mechanism that mimics the iterative nature of human code development.  A "code understanding" agent analyzes compiler feedback, identifying potential errors and generating targeted hints for improvement. These hints, along with positive and negative examples, are fed back to the code agents, driving them towards a more accurate and robust translation. This iterative refinement process continues until a satisfactory translation is achieved, or a predefined number of attempts is exhausted.

\subsection{Addressing the Nuances of Code}

Our architecture specifically addresses common pitfalls in code translation, such as violations of target language requirements, inappropriate duplication of source syntax, mismatches in API behaviors, and the unintentional removal of logic. Through a combination of knowledge grounding, concept-driven generation, and targeted feedback, \sln{} strives to produce translations that are not only functional but also adhere to the idiomatic conventions of the target language.


\begin{figure}
    \centering
    \includegraphics[width=.9\linewidth]{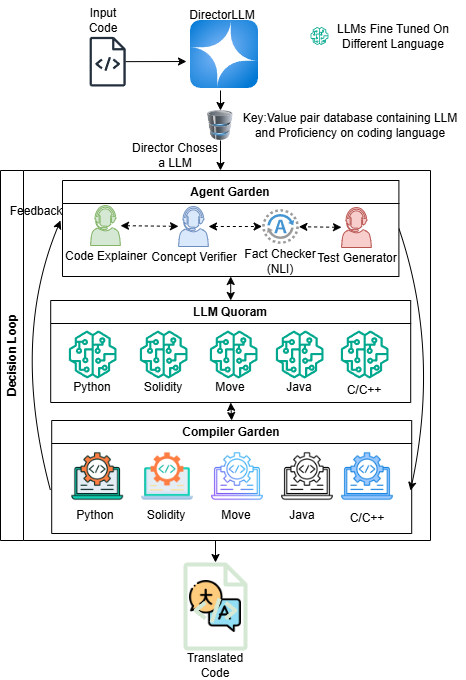}
    \caption{\sln{} Overall Workflow}
    \label{fig:workflow}
\end{figure}

\subsection{Workflow}
\sln{} workflow is depicted in high level at Figure~\ref{fig:workflow} multi-agent system architecture designed for robust code translation and optimization, leveraging a collection of specialized LLMs. The system initiates with the DirectorLLM, which orchestrates the process by analyzing the input code and selecting an appropriate LLM from the LLM Quorum – a pool of LLMs, each fine-tuned for a specific programming language (Python, Solidity, Move, Java, C/C++), based on a Key:Value pair database containing LLM proficiency benchmarks collected from respective model cards. The chosen LLM performs the initial code translation. Subsequently, the translated code enters a Decision Loop, where it's scrutinized by the Agent Garden – a set of specialized agents (Code Explainer, Concept Verifier, Fact Checker using Natural Language Inference, and Test Generator) that collaboratively analyze the code for correctness, consistency, and factual accuracy, and generate tests. Feedback from the Agent Garden, along with compilation results from the Compiler Garden (a set of compilers corresponding to the supported languages), is relayed back to the DirectorLLM, enabling it to either re-select a more suitable LLM or prompt the current LLM to refine the translation iteratively until convergence criteria are met, resulting in the final translated code.
\section{Preliminary Validation and Promising Results}
\label{sec:eval}

To demonstrate the potential of \sln{}, we conducted preliminary experiments focusing on challenging real-world code translation scenarios. We sourced code from a diverse set of GitHub projects, prioritizing those with substantial community recognition (measured by repository stars). Our evaluation included both widely used and low-resource languages, showcasing \sln{}'s versatility. Though \sln{} is model agnostic, we tested our model garden with granite-code, codegemma, deepseek-coder-v2, starcoder2, codegeex4, codestral, deepseek-coder, codellama, codeqwen, qwen2.5-coder, gemma2, gemma2, llama3.2, opencoder, llama3.3 models.

Our initial findings are highly encouraging. \sln{}, powered by a collaborative team of compact, open-source LLMs from the Gemma family \cite{team2024gemma}, achieved remarkable results, often rivaling and, in some cases, surpassing the performance of significantly larger, state-of-the-art models like GPT-4. This demonstrates the power of our multi-agent approach to unlock the capabilities of smaller, more accessible LLMs.

Table \ref{tab:transcoder} highlights the success rate of \sln{} in translating code across various language pairs, compared to other prominent models. Notably, \sln{} demonstrates strong performance in translating to and from languages like Go and Java, even outperforming GPT-4 in certain cases. These results underscore the effectiveness of our concept-guided translation and feedback-driven bug reduction strategies.

\begin{table}[]
\resizebox{\columnwidth}{!}{%
\begin{tabular}{|l|c|c|c|ccccc|}
\hline
\multirow{2}{*}{\textbf{Dataset}} &
  \multirow{3}{*}{Language} &
  \multirow{3}{*}{Samples} &
  \multirow{3}{*}{Language} &
  \multicolumn{1}{c|}{} &
  \multicolumn{1}{c|}{} &
  \multicolumn{1}{c|}{} &
  \multicolumn{1}{c|}{} &
   \\ \cline{5-9} 
 &
   &
   &
   &
  \multicolumn{1}{c|}{\textbf{CodeGeeX}} &
  \multicolumn{1}{c|}{\textbf{StarCoder}} &
  \multicolumn{1}{c|}{\textbf{GPT-4}} &
  \multicolumn{1}{c|}{\textbf{Llama 2}} &
  \textbf{TransCode} \\ \cline{1-1} \cline{5-9} 
\multirow{5}{*}{\cite{puri2codenet}} &
   &
   &
   &
  \multicolumn{5}{c|}{} \\ \cline{2-9} 
 &
  C++ &
  200 &
  C, Go, Java, Python &
  \multicolumn{1}{c|}{3.6\%} &
  \multicolumn{1}{c|}{39.1\%} &
  \multicolumn{1}{c|}{80.0\%} &
  \multicolumn{1}{c|}{9.5\%} &
  59.1\% \\ \cline{2-9} 
 &
  Go &
  200 &
  C, C++, Java, Python &
  \multicolumn{1}{c|}{5.9\%} &
  \multicolumn{1}{c|}{42.0\%} &
  \multicolumn{1}{c|}{85.5\%} &
  \multicolumn{1}{c|}{16.9\%} &
  76\% \\ \cline{2-9} 
 &
  Java &
  200 &
  C, C++, Go, Python &
  \multicolumn{1}{c|}{10.3\%} &
  \multicolumn{1}{c|}{30.3\%} &
  \multicolumn{1}{c|}{81.3\%} &
  \multicolumn{1}{c|}{13.9\%} &
  80.6\% \\ \cline{2-9} 
 &
  Python &
  200 &
  C, C++, Go, Java &
  \multicolumn{1}{c|}{7.3\%} &
  \multicolumn{1}{c|}{33.3\%} &
  \multicolumn{1}{c|}{79.9\%} &
  \multicolumn{1}{c|}{11.0\%} &
  71\% \\ \hline
\multirow{2}{*}{\cite{ahmad2021avatar}} &
  Java &
  249 &
  C, C++, Go, Python &
  \multicolumn{1}{c|}{1.8\%} &
  \multicolumn{1}{c|}{11.9\%} &
  \multicolumn{1}{c|}{70.8\%} &
  \multicolumn{1}{c|}{1.8\%} &
  71\% \\ \cline{2-9} 
 &
  Python &
  250 &
  C, C++, Go, Java &
  \multicolumn{1}{c|}{1.6\%} &
  \multicolumn{1}{c|}{14.2\%} &
  \multicolumn{1}{c|}{52.2\%} &
  \multicolumn{1}{c|}{4.7\%} &
  48\% \\ \hline
\end{tabular}%
}
\caption{Code Pair Translations Success Rates}
\label{tab:transcoder}
\end{table}

Furthermore, \sln{} exhibited a significant improvement in translating Solidity smart contracts to Move, a challenging task due to the unique characteristics of these languages. As shown in Table \ref{tab:solmover}, \sln{} successfully translated a substantial portion of the dataset, outperforming SolMover, a specialized tool for this specific language pair \cite{karanjai2024teaching}.

\begin{table}[h]
\centering
\caption{Solidity to Move Translation: \sln{} Shows Significant Improvement}
\label{tab:solmover}
\begin{tabular}{|l|c|c|}
\hline
\textbf{Model} & \textbf{Tasks Attempted} & \textbf{Success Rate} \\ \hline
\sln{} & 734 & \textbf{61.6\%} \\ \hline
SolMover & 734 & 47.5\% \\ \hline
\end{tabular}
\end{table}

While these results are based on initial experiments, they provide compelling evidence for the effectiveness of our approach. They highlight the potential of \sln{} to become a powerful tool for developers working across diverse programming languages, especially when computational resources are limited. Further development, including expanding language support and refining the agent interaction mechanisms, will unlock even greater capabilities.


Our experiments on the XLCoST benchmark \cite{zhu2022xlcost} provide further support for \sln{}'s capabilities (Table \ref{tab:xlcost}). While performance varied across language pairs, \sln{} demonstrated particularly strong results in translating to Python, surpassing existing models in several cases. This aligns with our vision of a versatile system capable of handling diverse translation tasks.


\begin{table}[]
\centering
\caption{CodeBLEU Scores on XLCoST: \sln{} Shows Competitive Performance}
\label{tab:xlcost}
\resizebox{\columnwidth}{!}{%
\begin{tabular}{|l|c|c|c|}
\hline
\textbf{Translation Task} &
  \textbf{\sln{}} &
  \textbf{SteloCoder~\cite{pan2023stelocoder}} &
  \textbf{XLCoST~\cite{zhu2022xlcost}} \\ \hline
C++ to Python        & \textbf{79.38} & 75.42          & 71.56 \\ \hline
C\# to Python        & 54.31          & \textbf{74.83} & 69.52 \\ \hline
JavaScript to Python & \textbf{80.07} & 73.05          & 68.42 \\ \hline
Java to Python       & 64.27          & \textbf{74.39} & 69.57 \\ \hline
PHP to Python        & 53.22          & \textbf{71.11} & 72.26 \\ \hline
\end{tabular}%
}
\end{table}

\section{Discussion: Charting the Future of Code Translation}
\label{sec:discussion}

Our initial exploration with \sln{} has yielded highly promising results, demonstrating that a collaborative framework of compact LLMs can achieve code translation performance that rivals, and at times surpasses, that of much larger models. This opens up exciting possibilities for democratizing access to advanced code translation capabilities, enabling deployment on a wider range of devices and in resource-constrained environments.

The success of our concept agent underscores the transformative potential of incorporating explicit programming knowledge into the translation process.  This approach not only improves accuracy but also paves the way for generating more semantically meaningful and idiomatic code, particularly for low-resource languages.  We envision extending this concept further, creating a bigger tool based framework that LLMs can leverage to understand the nuances of diverse programming paradigms.

Furthermore, the effectiveness of our feedback-driven bug reduction strategy highlights the importance of iterative refinement in code translation. By mimicking the way human developers use compiler feedback and debugging tools, \sln{} can proactively identify and address potential errors, leading to more robust and reliable translations.

\sln{}'s modular design also makes it a versatile platform for future innovation.  While our initial experiments focused on a specific family of LLMs, the framework can readily accommodate other models, including larger, more powerful ones. This flexibility allows for continuous improvement and adaptation as new and more capable LLMs emerge.

Looking ahead, we see several exciting avenues for future work:

\begin{itemize}
    \item \textbf{Expanding Language Coverage:} We aim to significantly expand the range of supported languages, with a particular focus on bridging the gap between mainstream and low-resource languages.
    \item \textbf{Enhancing Agent Specialization:} We will explore more sophisticated methods for training and deploying specialized agents, each a master of a specific programming language, paradigm, or even a particular aspect of the translation process.
    \item \textbf{Developing Adaptive Learning Mechanisms:}  We envision \sln{} as a continuous learning system capable of adapting to new code patterns, evolving language standards, and user feedback.
    \item \textbf{Integrating with Developer Workflows:} We plan to explore seamless integration with existing development tools and workflows, making \sln{} an indispensable asset for programmers worldwide.
    \item \textbf{Model+Agent Recipe Recommendation:} One avenue we identified is, having a more robust and exhaustive benchmark with various LLMs and agent combinations will allow us to come up with weighted recommendation on which combinations work better and will open a avenue for further optimization research.
\end{itemize}


\section*{Limitations}
\label{sec:limitations}

While \sln{} shows very early promising results, it is important to acknowledge its current limitations. The performance of smaller LLMs may not always match that of larger models, particularly in complex scenarios. Potential biases and hallucinations in LLMs can lead to errors, and the framework's performance may vary across different programming languages and domains. Addressing these limitations, alongside potential security concerns from automated code modification, will require ongoing research and robust mitigation strategies. We need more work on our NLI system and a more exhaustive research to make the framework more robust and find its more nuanced pitfalls.

\section{Conclusion}
\label{sec:conclusion}

\sln{} represents a paradigm shift in code translation. By embracing a collaborative, multi-agent framework guided by conceptual knowledge, we have demonstrated that even compact LLMs can achieve remarkable accuracy and efficiency. This research not only challenges the prevalent wisdom that "bigger is always better" but also unlocks the potential agentic framework by having a pipeline that can dynamically create code translation recipes based on available LLMs and their strengths, virtually creating a huge number of translation pathways..


The implications of this vision extend far beyond the realm of code translation.  We believe that the principles embodied in \sln{} – collaboration, specialization, and knowledge grounding – can serve as a blueprint for a new generation of AI systems that are more efficient, adaptable, and ultimately, more human-centric.  As we continue to refine and expand \sln{}, we are confident that it will play a pivotal role in shaping a future where code flows freely, empowering developers to build a more interconnected and innovative world.

\section*{Acknowledgments}
This research was supported by the Sui Foundation Academic Grant (SARA), awarded to Rabimba Karanjai and Weidong Shi. Additionally, this work benefited from the generous provision of compute credits through the Google Developer Expert (GDE) and Google Cloud Research Innovator program, with special thanks to the GDE AI team.

\bibliographystyle{ACM-Reference-Format}
\bibliography{sample-base}
\end{document}